\documentclass[preprint, 10pt]{elsarticle}

\usepackage{amssymb}
\usepackage{amsmath}

\usepackage{lineno}

\usepackage{graphicx}
\usepackage{epstopdf}
\usepackage{amsmath}
\usepackage{amssymb}
\usepackage{booktabs}
\usepackage{wrapfig}
\usepackage{multirow}
\usepackage{stfloats}
\usepackage{float}
\usepackage{hyperref}
\usepackage[capitalize]{cleveref}
\crefname{section}{Sec.}{Secs.}
\Crefname{section}{Section}{Sections}
\Crefname{table}{Table}{Tables}
\crefname{table}{Tab.}{Tabs.}

\newcommand{\ie }{\emph{i.e.}}

\newcommand{\etc}{etc}

\usepackage[capitalize]{cleveref}
\crefname{section}{Sec.}{Secs.}
\Crefname{section}{Section}{Sections}
\Crefname{table}{Table}{Tables}
\crefname{table}{Tab.}{Tabs.}
\usepackage{graphicx}
\usepackage{wrapfig}
\usepackage{xspace}
\usepackage{multirow}
\usepackage{colortbl}
\usepackage{color}
\usepackage{caption}
\usepackage{subcaption}
\usepackage{wrapfig}
\usepackage{enumitem}
\usepackage{makecell}
\usepackage{tabularx}
\usepackage{svg}
\usepackage{stfloats}
\makeatletter
\DeclareRobustCommand\onedot{\futurelet\@let@token\@onedot}
\def\@onedot{\ifx\@let@token.\else.\null\fi\xspace}

\def\ie{\emph{i.e}\onedot} 
 
\def\etc{\emph{etc}\onedot}

\makeatother

\usepackage{float} 
\usepackage{booktabs}
\begin{document}

\begin{frontmatter}



\title{ChatSearch: a Dataset and a Generative Retrieval Model for General Conversational Image Retrieval}

\author{Zijia Zhao\fnref{lab,school}}
\ead{zhaozijia2021@ia.ac.cn}
\author{Longteng Guo\fnref{lab}}
\ead{longteng.guo@nlpr.ia.ac.cn}
\author{Tongtian Yue\fnref{lab,school}}
\ead{yuetongtian2022@ia.ac.cn}
\author{Erdong Hu\fnref{lab,school}}
\ead{huerdong2022@ia.ac.cn}
\author{Shuai Shao\fnref{company}}
\ead{shaoshuai@acm.org} 

\author{Zehuan Yuan\fnref{company}}
\ead{yuanzehuan@bytedance.com}
\author{Hua Huang\fnref{bnu}}
\ead{huahuang@bnu.edu.cn}
\author{Jing Liu\corref{cor1}\fnref{lab,school}}
\ead{jliu@nlpr.ia.ac.cn}
\cortext[cor1]{Corresponding author.}
\fntext[lab]{The Laboratory of Cognition and Decision Intelligence for Complex Systems, Institute of Automation, Chinese Academy of Sciences, Beijing, 100190, China.}
\fntext[school]{School of Artificial Intelligence, University of Chinese Academy of Sciences, Beijing, 100049, China}
\fntext[company]{Bytedance Inc., Beijing, 100098, China}
\fntext[bnu]{School of Artificial Intelligence, Beijing Normal University, Beijing, 102211, China}


\begin{abstract}
In this paper, we investigate the task of general conversational image retrieval on open-domain images.
The objective is to search for images based on interactive conversations between humans and computers. To advance this task, we curate a dataset called ChatSearch. This dataset includes a \textbf{multi-round multimodal conversational context query} for each target image, thereby requiring the retrieval system to find the accurate image from database. 
Simultaneously, we propose a generative retrieval model named ChatSearcher, which is trained end-to-end to accept/produce interleaved image-text inputs/outputs. 
ChatSearcher exhibits strong capability in reasoning with multimodal context and can leverage world knowledge to yield visual retrieval results. 
It demonstrates superior performance on the ChatSearch dataset and also achieves competitive results on other image retrieval tasks and visual conversation tasks. We anticipate that this work will inspire further research on interactive multimodal retrieval systems. Our dataset will be available at \url{https://github.com/joez17/ChatSearch}.
\end{abstract}




\end{frontmatter}

\date{}

\section{Introduction}
\label{sec:intro}
Image retrieval is a task that focuses on searching for desired images corresponding to an abstract concept formed in a user's mind, where the user need to somehow convey this concept through human-computer interaction.
Various interaction methods have been investigated for image retrieval.
One naive interaction approach involves using content related to the desired image, such as a reference image~\cite{tong2001supportir}, a set of attributes~\cite{felix2012weak} or a descriptive sentence~\cite{li2011textir, clip}. 
Enhanced interaction methods are employed to refine image retrieval results by incorporating strategies like relevance score~\cite{rui1998relevance} or textual user feedback~\cite{liu2021cirr, guo2018dialogir, yuan2021conversational}.

The emergence of ChatGPT has shown that a disarmingly simple conversation can serve as an ideal interaction interface for a powerful natural language generation system. 
Such type of conversation is also an intuitive and effective interaction interface for image retrieval system, offering the following advantages: 
1) \textbf{Natural interaction}: Users are able to engage with the system using their preferred communication style, resulting in a more natural and seamless retrieval process. 
2) \textbf{Comprehensive context}: Through conversation, the retrieval system can comprehend and consider the overall context, thereby providing users with more relevant and precise information.
3) \textbf{Interactive feedback}: In multi-round conversation, the retrieval results can be progressively refined through interactive user feedback.
4) \textbf{Multimodal experience}: Both the user and the system can gather and express information through both image and text, creating a integrated multimodal interaction experience.  

\begin{figure*}[t!]
  \centering
     \includegraphics[width=\linewidth]{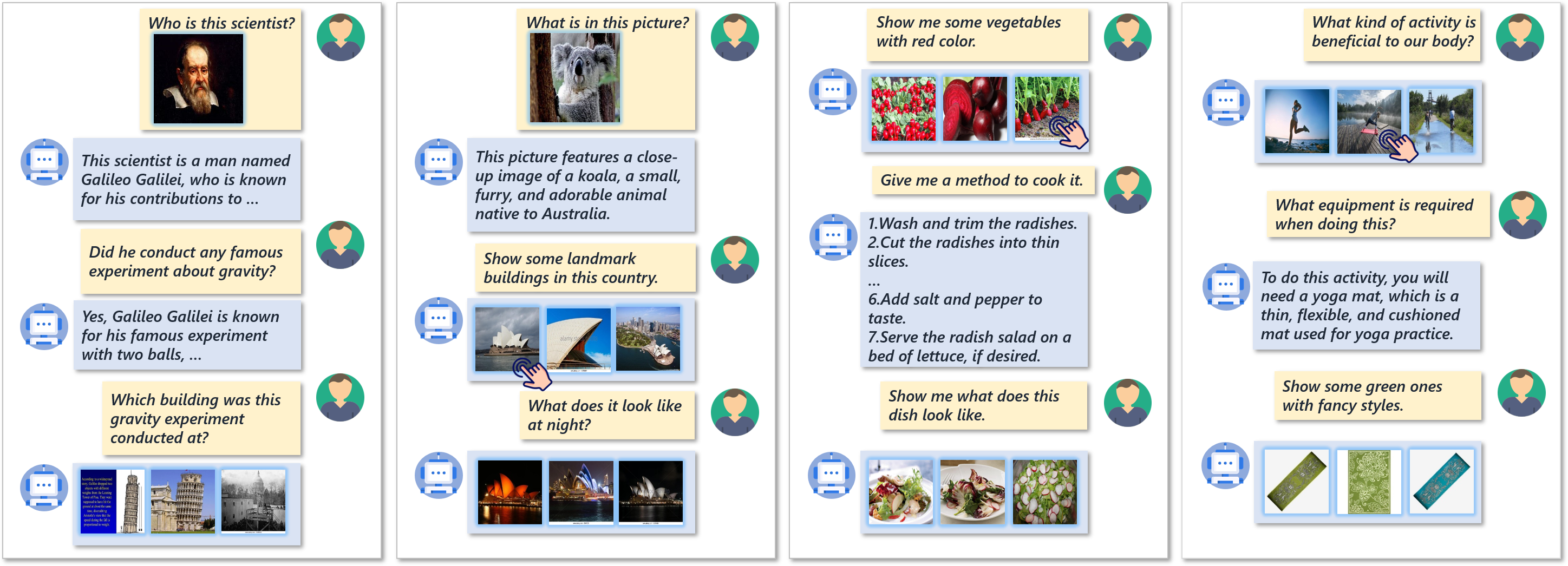}
  \caption{Our generative retrieval model ChatSearcher can accept multimodal inputs and generate textual response or retrieved images. ChatSearcher can comprehend multimodal dialogue context, infer user's implicit intentions, generate visual or textual responses through multimodal reasoning and world knowledge, and can also support interactive refinement of results.}
\label{ills}
\end{figure*}
In this paper, we investigate the task of general conversational image retrieval on open-domain images. To advance this task, we create a dataset called \textbf{ChatSearch}, which necessitates the retrieval model to search for desired images by perceiving a \textbf{multi-round multimodal dialogue} containing both textual and visual human-computer interactions. 
In ChatSearch, the information needed to retrieve the images is not explicitly stated but often implied within the context of the dialogue. This necessitates the retrieval model to acquire such information through multimodal comprehension, complex reasoning, and world knowledge.  
To construct ChatSearch, we initially employ a meticulously designed automatic pipeline with the assistance of large-scale pretrained models. Subsequently, the dataset undergoes a manual review process conducted by domain experts.

We also introduce a generative retrieval model called \textbf{ChatSearcher} specifically designed for conversational image retrieval.
ChatSearcher is end-to-end trained to accept interleaved image-text inputs and produce relevant outputs that also combine both images and text in an interleaved format. 
To accomplish this, we leverage the advanced capabilities of a large language model (LLM).
We extract visual embeddings for images in the interleaved input sequence, and concatenate them with textual tokens to form a multimodal token sequence.
Specifically, we employ a unified-format training objective for the multimodal sequence, treating both word prediction and image retrieval as generative progresses. In word prediction, we optimize for the probability of ground-truth word prediction within the word vocabulary. For image retrieval, we maximize the probability of image feature matching within a dynamically updated image feature queue, which can be viewed as a visual vocabulary. 
The training of ChatSearcher involves a two-stage procedure: establishing bidirectional image-text alignment using interleaved image-text data and conversational instruction tuning with diverse instruction data. 
This instruction data includes conversational image retrieval instructions, visual conversation instructions, and instructions for manipulating AI-generated content (AIGC) images. 
The derived ChatSearcher model can effectively reasoning out the retrieval query embedding from complex multimodal dialogue context and perform relevance ranking to retrieve the desired images.

Our paper makes the following contributions:
\begin{itemize}
    \item We introduce ChatSearch, a dataset for general conversational image retrieval. This dataset emphasizes the need for multimodal reasoning based on multi-round conversations, which is essential for building an intuitive interaction interface for intelligent retrieval systems. 
    \item We propose ChatSearcher, a generative retrieval model that is trained end-to-end to accept and produce interleaved image-text inputs/outputs. 
    \item ChatSearcher demonstrates superior performance on the general conversational image retrieval task. Additionally, it exhibits strong generalization capabilities to other image retrieval tasks, maintaining competitive performances on zero-shot text-to-image retrieval and zero-shot composed image retrieval. It also shows comparable performace on visual conversation tasks.

\end{itemize}

\section{Related Works}
\subsection{Image Retrieval}
Image retrieval has been a widely researched topic. Traditional content-based image retrieval~\cite{tong2001supportir, liu2011imagepr, huang2003imagepr, mahmoudi2003imagepr, liu2013contentpr} tasks use an image as a query to identify the desired image. Cross-modal image retrieval introduces external query modalities, such as text-to-image retrieval~\cite{wang2016learning,clip} or sketch-to-image retrieval~\cite{sangkloy2016sketchy}, \etc.
Interactive image retrieval is raised to refine the image retrieval results with the help of human-computer interaction. Early methods uses simple user feedback to interact with the retrieval system, including relevance~\cite{rui1998relevance} and attribute~\cite{ak2018learning}. Some advanced works propose to adopt natural language user feedback to interact~\cite{guo2018dialogir, liu2021cirr}, which is more familiar with human users. Some works study the image retrieval tasks based on a simple-format human-computer conversation in special area~\cite{ yuan2021conversational}. 
However, these studies primarily focus on single round interaction or be limited in some special domains like fashion images.
In this work, we propose general conversational image retrieval task, aiming to execute image retrieval predicated on an advanced and adaptable form of multimodal conversation in a wider open-domain setting. This necessitates the model to comprehend both general visual and textual context, accommodate the intrinsic user intentions, demanding complex reasoning and the invocation of world knowledge.

\subsection{Multimodal Large Language Models}
With the emergence of ChatGPT, human-computer conversational interactions have become a focal point of contemporary discourse. Delving deeper, some researchers have embarked on integrating visual content into dialogues with a Multimodal Large Language Model~\cite{blip2,llava,liu2023improved}. 
Recently, studies have explored the intersection between multimodal LLMs and multimodal tasks, including multi-task learning~\cite{dai2023instructblip}, multimodal in-context learning~\cite{laurenccon2023introducing}, multimodal output ~\cite{koh2023fromage}, \etc.

\section{ChatSearch: A General Conversational Image Retrieval Dataset}
\label{sec:formatting}
In this paper, we study a complex situation: retrieving the image based on a multi-round multimodal conversation between human and retrieval system. This task requires the retrieval system to comprehend multimodal contents, as well as extract the retrieval intention from multi-round dialogues.
Owing to the absence of existing datasets, we construct a general conversational image retrieval dataset ChatSearch with real-world images. 
\subsection{Automatic Construction Pipeline}
\label{datapipeline}
\begin{figure}[t!]
\centering

\includegraphics[width=0.7\linewidth]{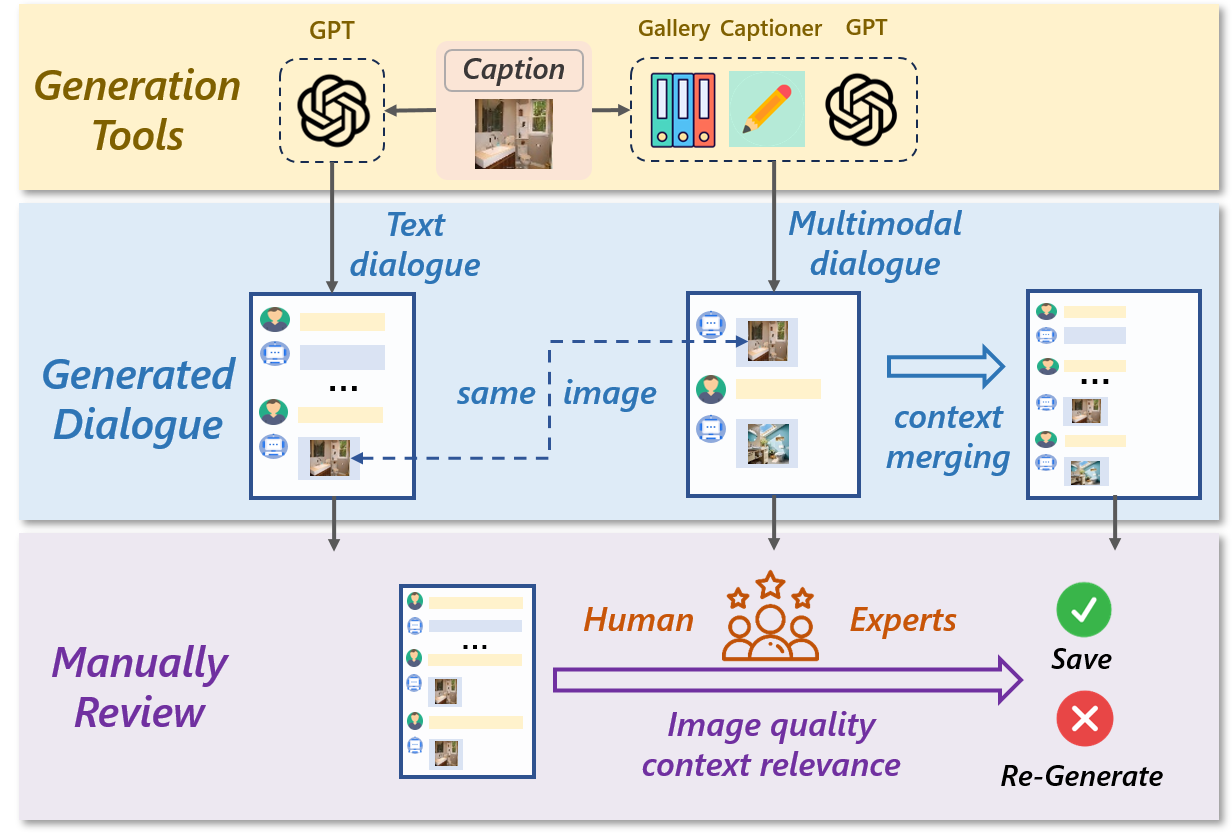}
\caption{Illustration of automatic data construction pipeline for general conversational image retrieval dataset ChatSearch. We use foundation models (text generator GPT, image gallery retriever and image captioner) as generation tools to generate \textit{text dialogue} and \textit{multimodal dialogue} that aim at searching user desired images, as elaborated in \cref{refgen}. Then we apply \textit{context merging} method and manually review on those data to construct a high quality evaluation split.}
\label{merge}
\end{figure}
The whole automatic data construction pipeline is shown in \cref{merge}. 
Our target is to construct a multimodal dialogue context to search for an image in corpus.
During the specific construction process, we primarily expand the existing real-world image-text retrieval dataset MSCOCO~\cite{coco}. 
We utilized existed foundation models including a text generator GPT-4~\cite{gpt4}, a gallery retriever CLIP-H~\cite{clip} and a pre-trained image captioner BLIP-2-OPT2.7b~\cite{blip2} to assist us in constructing dialogues for image retrieval, instead of relying on external human annotation. After dialogue construction, we apply context merging method to get more complex multimodal dialogues. Finally, we perform manually review on evaluation split data according to the image quality and context relevance in generated dialogues.

\paragraph{Text dialogue context construction} We select one raw image caption from the image-text pair in MSCOCO, send it to the text generator GPT-4 and require the text generator to generate a multi-round textual dialogue query for image retrieval with special prompts.
The model needs to reason on the whole dialogue's information and invoke outside-world knowledge to get the correct image. 

\begin{figure*}[t!]
\centering
\subfloat[{Dialogue construction with reference image (MDC-I). We use the image $I_1$ from source image-text pair as the \textit{reference image} to generate a single-round dialogue for image retrieval.}]{
		\includegraphics[width=0.47\linewidth]{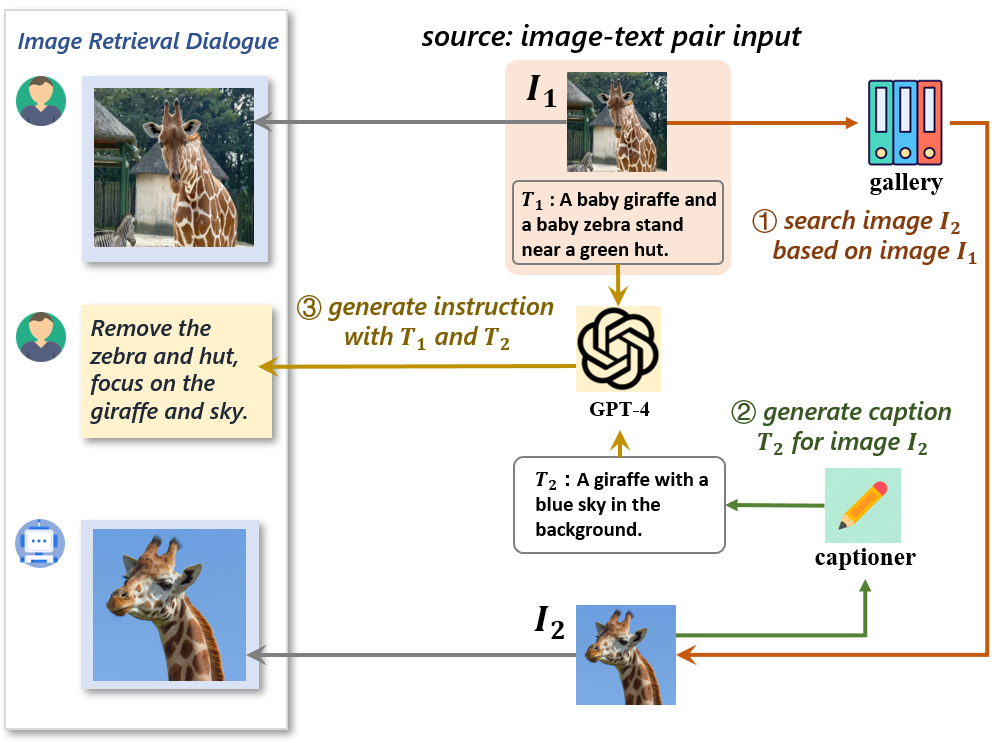}
  \label{refimage}}
  \hspace{0.01\linewidth}
\subfloat[{Dialogue construction with reference text (MDC-T). We use the text $T_1$ from source image-text pair as the \textit{reference text} to generate 2-round multimodal dialogue for image retrieval.}]{
		\includegraphics[width=0.47\linewidth]{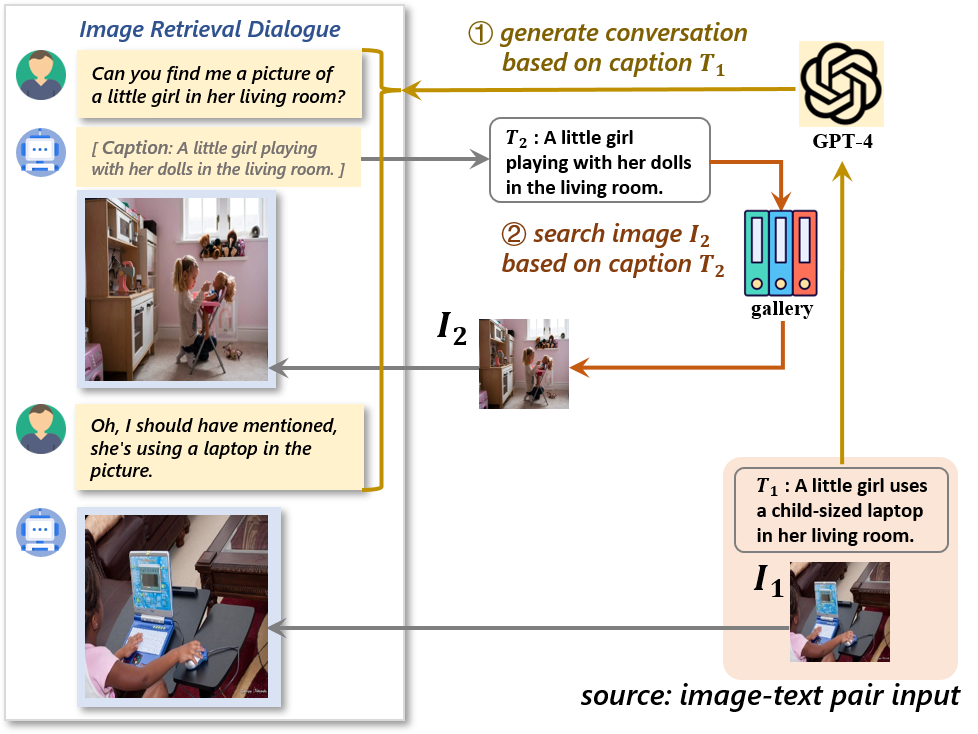}
  \label{reftext}}
\caption{{Multimodal dialogue construction. The whole pipeline is combined with a text generator, a image search gallery and a pre-trained image captioner. The final output is a user-assistant multimodal conversation designed for image retrieval. } }

\label{refgen}
\end{figure*}

\paragraph{Multimodal dialogue context construction}
In this part, we generate multimodal dialogue data for image retrieval, which contains both visual and textual contents in the conversation.
These kind of data require the model to understand the textual user instructions and comprehend the image context as well. We use image-caption pair from MSCOCO as reference image and text to construct dialogues. The detailed pipelines are shown in \cref{refgen}.
\noindent
\begin{itemize}[leftmargin=0.5cm, itemindent=0cm]
\item \textbf{Dialogue construction with reference image (MDC-I).} We generate a single-round multimodal dialogue that includes a user query containing both visual and textual content, along with a target image to be retrieved based on the user query, as shown in ~\cref{refimage}. The dialogue is constructed using a reference image $I_1$ from the MSCOCO image-text pair $(I_1, T_1)$. First, we use the CLIP-H model to search for the top 10 similar images from the LAION-5B gallery~\cite{laion5b} based on $I_1$. After applying a similarity threshold, one image $I_2$ is randomly selected from the remaining retrieved images. A pre-trained image captioner then generates a pseudo target caption $T_2$ for $I_2$. Finally, the source caption $T_1$ and target caption $T_2$ are input into GPT-4 to produce user modification instructions, with constraints applied to emphasize the primary differences between the two images.
\item \textbf{Dialogue construction with reference text (MDC-T).} We generate a two-round image-text dialogue containing two rounds of textual user query and assistant-given target image in ~\cref{reftext}. The dialogue is constructed based on a reference text $T_1$ from the source image-text pair $(I_1, T_1)$ from MSCOCO.
We first input the reference text $T_1$ into GPT-4 to generate a 2-round dialogue, where the answer in the first round should be a caption $T_2$ representing the visual context, and the final answer is the given target caption $T_1$. We then extract the caption $T_2$ from the first round's response. Using the CLIP-H model, we search for the most relevant image $I_2$ based on $T_2$ from the LAION-5B gallery~\cite{laion5b}. Finally, we replace the extracted caption $T_2$ with the retrieved image $I_2$ to construct a 2-round image-text dialogue. Constraints are applied to GPT-4 to maintain the semantic connection between the two images.
\end{itemize}
The above reference image and reference text methods bring different natures of the source and target images pair in the constructed dialogue. In dialogues constructed with reference image (MDC-I), the image pair mostly share the same subject, with different background or orientations. In contrary, the image pair in dialogues constructed with reference text (MDC-T) have more difference in attributes, varieties and quantities. 
This result can be attributed to the nature of CLIP embeddings, which primarily emphasize the main object within an image, whereas caption modifications via GPT-4 is more free-form and diverse. 
Thus, combining these two methods enrich the diversity of the constructed data.

\paragraph{Dialogue construction with context merging}
After the generation of above data, we merge the plain textual dialogue and the single-round image-text dialogue which has common images together to construct complex interactive dialogues in \cref{merge}, 
providing a more challenging image retrieval task for retrieval systems. 
\begin{table}[t!]
\centering

\resizebox{0.9\linewidth}{!}{
\begin{tabular}{lccc}
\toprule
          & tChatSearch & iChatSearch&mChatSearch \\
          \midrule

Context modality      & text    & image-text       & image-text        \\    \midrule
Sample number & 5000    &    10000           &     10000   \\    \midrule
Context length & 66.7    &   12.4          &     56.5   \\    \midrule
{Task input} & \makecell[c]{{multi-round}\\ {textual dialogue}}    &   \makecell[c]{{single image}\\ { and text instruction}}         &  \makecell[c]{{multi-round}\\ {multimodal dialogue}  }\\  \midrule
{Task output} & {image candidates}    &  {image candidates}    & {image candidates }\\
\bottomrule
\end{tabular}}
\label{statistic}
\caption{Statistics of ChatSearch test split.}
\end{table}

\subsection{Benchmark}
We select the data sourced from test and val karpathy~\cite{karpathy} split of MSCOCO to compose the ChatSearch test split, while the others are utilized for training.
We let five human experts to manually review these test split data and re-generate the unqualified one by following rules.
1) Image quality: Ensuring the retrieved images from LAION-5B are clear, concise, and harmless. 
2) Context relevance: Checking whether the multimodal content in the generated dialogue has a reasonable logical and relevance relationship.

As shown in~\cref{statistic}, we divided ChatSearch into three sub-tasks: tChatSearch, iChatSearch and mChatSearch, according to the format of dialogue context. tChatSearch is based on the multi-round plain text dialogue context. iChatSearch is based on a single-round image-text context, including reference-image data and reference-text data. mChatSearch is based on a multimodal multi-round dialogue context, which contains complex merged dialogue context and 2-round image-text dialogue generated by reference text. 

All sub-tasks are evaluated with recall rate at rank 1, 5, 10. Recall at rank $K$ (R$@K$) quantifies the number of times the correct image is among the top $K$ results. Higher recall means better performance. And we also compute the average recall rate to reflect a general ability on conversational image retrieval.

\section{ChatSearcher: A Generative Retrieval Model}
We introduce ChatSearcher, a generative model that is trained end-to-end to accept interleaved image-text inputs and produce relevant outputs that also combine both retrieved images and generated text in an interleaved format.
\begin{figure*}[t]
  \centering
     \includegraphics[width=\linewidth]{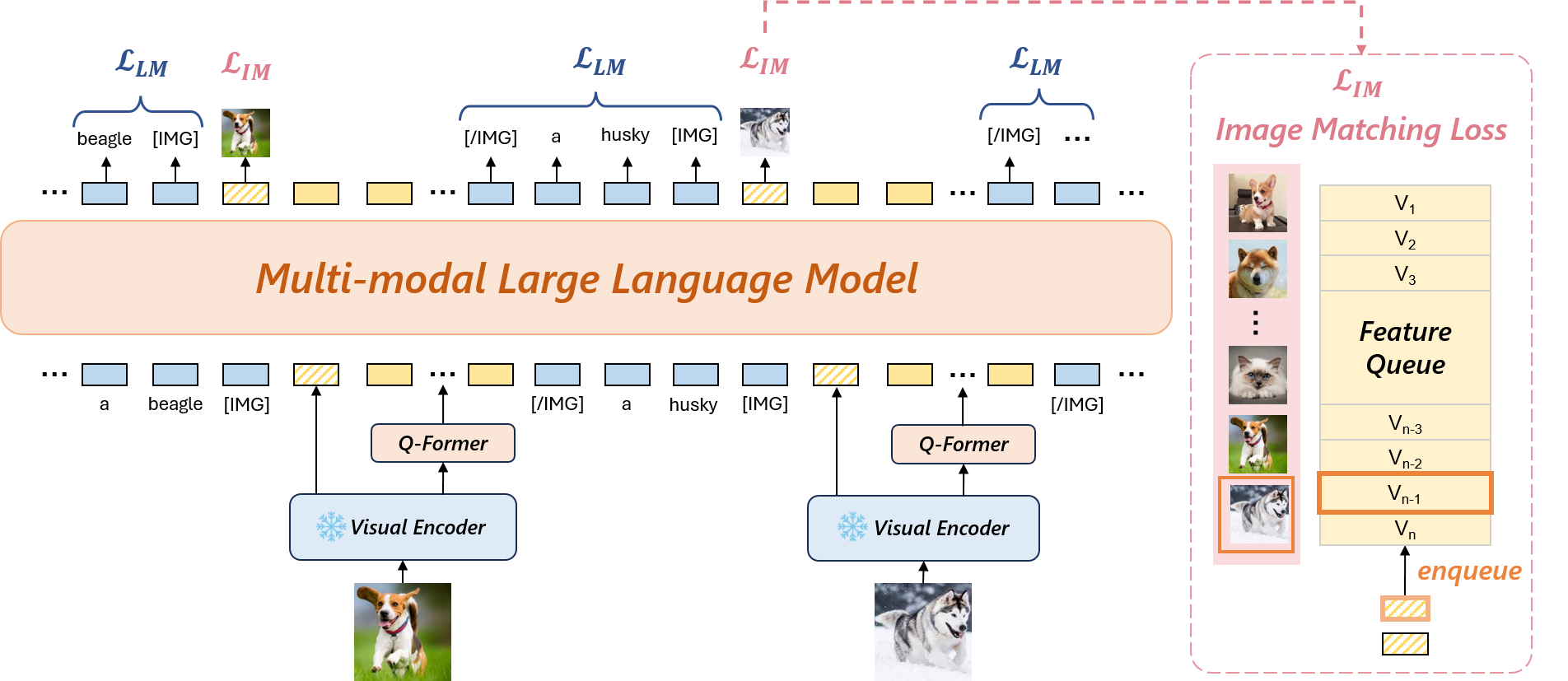}
  \caption{Architecture of our generative retrieval model ChatSearcher. Interleaved documents serve as input, predicting words or retrieving images with generative training objective. Special token $\mathtt{[IMG]}$ predicts where to retrieve images. We use a dynamicaly-updated feature queue to save contrastive samples for image retrieval. }
  
\label{arch}
\end{figure*}
\subsection{Architecture}
As shown in~\cref{arch}, our model is built with a causal decoder-only LLM, which is initialized with Vicuna-7B v1.5~\cite{vicuna2023}.
We use OpenAI's CLIP VIT-L~\cite{clip} as the vision backbone to generate visual encodings $f$ and global feature $f_{\mathtt{\langle CLS \rangle}}$ for each input image. 
Then we use a Q-former perceiver to compress the visual encodings $f$ from the vision backbone into a fixed number of dense embeddings $q$. The perceiver is initialized with the weight of BLIP-2 Q-former~\cite{blip2}. 
And we use different linear projectors to project global visual feature $f_{\mathtt{\langle CLS \rangle}}$ and dense embeddings $q$ into the same dimensions.
Then we concatenate them with two special tokens $\mathtt{[IMG]}$ and $\mathtt{[/ IMG]}$ as the visual inputs of the LLM decoder:
$\mathtt{[IMG] \langle CLS \rangle \langle q_1 \rangle \langle q_2 \rangle ... \langle q_N \rangle [/ IMG]}$.  
We extract the above visual embeddings for each image in the interleaved image-text document and combine them with other textual tokens. 
We use independent linear head $\mathcal{M}_t$ and $\mathcal{M}_v$ on text output and visual output for text generation and image retrieval respectively.

To enhance the retrieval ability with numerous negative image samples, we build a large sample dictionary that covers a rich set of sample features, which are the global visual feature $f_{\mathtt{\langle CLS \rangle}}$ extracted from the frozen vision encoder. 
We use a dynamically updated feature queue $\mathcal{Q}$ as the sample dictionary inspired by \cite{wu2018memorybank}. 
The samples in the dictionary are progressively replaced by enqueuing the image features from the current mini-batch and removing the oldest mini-batch. 
\subsection{Generative Training Objective}
\label{gen_obj}

We use a unified-format training objective for multimodal input sequence, regarding word prediction and image retrieval both as generative progresses.
Given an multimodal token sequence $w=\{w_t\}^T_{t=1}$ which contains text tokens and visual tokens, we maximize the likelihood of the ground-truth token:
    \begin{equation}
        \mathcal{L}(w) = \sum_{w_t\in \mathcal{D}} \mathrm{log}P(w_t|w_{<t};\theta, \epsilon, \mathcal{Q})
    \end{equation}
    
where $\mathcal{D}$ is composed with the multimodal vocabulary except the Q-former feature tokens $\mathtt{\langle q_{1...N} \rangle}$. 

The conditional probability $P$ is modeled respectively for image and text tokens in sequence $w$.
For text tokens in the sequence, we use the standard language modeling objective, computing the probability that the predicted word is identical to the correct word.
For image in the sequence, the model will generate an image retrieval embedding $h_I$ after being queried with the special $\mathtt{[IMG]}$ token. Then, we use a linear projector $\mathcal{M}_v$ to project it into the retrieval embedding space $\Phi$. 
And we use a linear projector $\mathcal{M_Q}$ to project the features in queue $\mathcal{Q}$ into the retrieval embedding space $\Phi$ as well. The normalized cosine similarity for the retrieval query $x$ and sample $y$ in the queue can be computed as:
\begin{equation}
    \mathrm{sim}(x,y) = \frac{(x^T\mathcal{M}_v)(y^T\mathcal{M_Q})^T}{\Vert x^T\mathcal{M}_v \Vert \Vert y^T\mathcal{M_Q} \Vert}
\end{equation}

And we compute feature matching conditional probability with the queue samples for all image samples in the sequence:
\begin{equation}
            P(w_I|w_{<I};\theta, \epsilon, \mathcal{Q}) = \frac{\mathrm{exp}(\mathrm{sim}(h_I,\mathcal{Q}_{pos})/\tau)}{\sum_{j=1}^{|\mathcal{Q}|}\mathrm{exp}(\mathrm{sim}(h_I,\mathcal{Q}_{j})/\tau)}
\end{equation}
where $\tau$ is the learnable temperature parameter and $I$ represents the index of $\mathtt{\langle CLS \rangle}$ token in the multimodal sequence $w$.
\subsection{Training Details}
We use a two-stage training strategy to train our model. 
\paragraph{Stage1: Bidirectional Image-Text Alignment} In the first stage, we try to build the bidirectional image-text alignment (\ie both image-to-text generation and text-to-image retrieval are learned) in our model using interleaved image-text data. We use CC3M~\cite{cc3m} and mmc4-core~\cite{zhu2023mmc4} dataset to pretrain our model. For CC3M image-text pair dataset, we arranged the images and captions in two different configurations: before and after the caption. For mmc4-core interleaved image-text document dataset, we randomly place the image before or after its corresponding sentence and divide the overly long multimodal sequences into several smaller segments at a fixed length for highly efficient training.  
In this stage, we train the model for 200k steps using a batch size of 256 for CC3M data and 64 for mmc4-core data on 8 NVIDIA A100 GPUs. We use the
AdamW~\cite{adamw}  optimizer with a weight decay of 0. The learning rate is warmed-up to $2\times10^{-5}$ in the first 6k iterations following a cosine schedule. We use $224\times224$ image resolution for all images. The queue size of negative sample dictionary is set to 1000.
\paragraph{Stage2: Conversational Instruction Tuning} 
To help the model better follow the different multimodal instructions,
we construct an instruction tuning dataset with LLaVA-150k~\cite{llava}, 10k samples from InstructPix2Pix~\cite{brooks2023instructpix2pix} and 10k samples from ChatSearch training set. 
LLaVA-150k contains different kind of visual conversation with textual response, while InstructPix2Pix and ChatSearch contains images in dialogue's response, which can be formatted as conversational image retrieval instructions furthermore. 
 Since Instructpix2pix~\cite{brooks2023instructpix2pix} is an image editing dataset which contains a triplet of reference image, editing instruction and target image, we simply modify the editing instruction into a more human-like textual instruction with some pre-defined template. Then we can construct a single-round conversation in which user provide an reference image and an editing instruction to require the model to find the target image with retrieval. 
Then, these three kind of datasets are mixed to perform conversational instruction tuning.
We employ a question-answer template like ``$\mathtt{USER:} \langle question \rangle \ \mathtt{ASSISTANT:} \langle answer \rangle$" to unified the instruction format and use a \texttt{[image]} placeholder to represent the image content in the conversation.
We compute the text and image loss using generative training objective proposed in \cref{gen_obj} on assistant's answers in each round of the instruction dialogues.
In this stage, we train the model for 10k steps using a batch size of 64 on 8 NVIDIA A100 GPUs. We use the AdamW~\cite{adamw}  optimizer with a weight decay of 0. The learning rate is warmed-up to $2\times10^{-6}$ in the first 300 iterations following a cosine schedule. We use $224\times224$ image resolution for all images. The queue size of negative sample dictionary is set to 1000.

\subsection{Interactive Inference}
User can interact with ChatSearcher in two ways: providing multimodal instructions or selecting from candidate image results. 
The model automatically determines whether to output a retrieved image by producing the special $\mathtt{[IMG]}$ token based on the multimodal dialogue context.
Utilizing the feature embedding of this special query token, the model outputs a set of image candidates ordered by feature similarity. 
The user can select one of these candidates to continue the interaction. Meanwhile, the selected image is appended to the end of the historical conversation sequence for continuous generation.

\section{Experiments}
\begin{table*}[t!]
\centering

\resizebox{\linewidth}{!}{
\begin{tabular}{l|ccc|ccc|ccc|c}
\toprule
    \multirow{2}{*}{Method}     & \multicolumn{3}{c|}{tChatSearch}   & \multicolumn{3}{c|}{iChatSearch}             & \multicolumn{3}{c|}{mChatSearch}  &  \multirow{2}{*}{Avg.}         \\
              & R@1&R@5&R@10 & R@1&R@5&R@10 & R@1&R@5&R@10& \\
\midrule
random choice &  0.02&0.1&0.2   &  0.01&0.05&0.1   &   0.01&0.05&0.1 &0.07  \\
CLIP-i~\cite{clip}    &  -&-&-&    9.65&20.96&28.05&  9.65&20.96&28.05 & 19.55\\
CLIP-t~\cite{clip}       &  15.84&34.46&45.10&  14.15&30.60&39.56& 12.69&27.19&35.80&28.38\\

CLIP-ti~\cite{clip}     &-  & -&-&  12.33  & 26.81 & 35.15    & 13.18  & 28.56  & 37.07&25.52\\
FROMAGe~\cite{koh2023fromage}       &  15.94 &  36.76  &  48.60 &  12.56 & 29.65 & 39.65    & 14.36  & 32.58  & 42.80&30.32\\
\midrule
ChatSearcher &  \textbf{27.38} & \textbf{52.48}& \textbf{63.50} & \textbf{35.54}  & \textbf{61.16} & \textbf{71.57}& \textbf{37.90} & \textbf{64.22}  &  \textbf{74.06}&\textbf{54.20} \\
\bottomrule
\end{tabular}}

\label{chatsearch}
\caption{General conversational image retrieval results on ChatSearch test split.}
\end{table*}

%

\subsection{General Conversational Image Retrieval}
\label{clip-it}


We evaluate the general conversational image retrieval performance on ChatSearch test split across three distinct tasks: tChatSearch (searching for image based on multi-round textual conversation), iChatSearch (searching for image based on single image and textual instruction) and mChatSearch (searching for image based on multi-round multimodal conversation). 

We compare our model against the baseline model CLIP~\cite{clip} and a Multimodal LLM model FROMAGe~\cite{koh2023fromage}. Given that the CLIP model exclusively accepts either image or text for feature extraction, the performance of CLIP is presented under three configurations: retrieval of images via dialogue text (CLIP-t), retrieval of images through dialogue image features (CLIP-i), and retrieval of images by amalgamating both dialogue text and dialogue image features (CLIP-ti). 

As shown in \cref{t2i}, CLIP shows strong ability in traditional text-based image retrieval. However, it fails in general conversational image retrieval tasks according to the results in \cref{chatsearch}. This indicates that while some traditional retrieval models can understand explicit textual expression, they remain limited in comprehending multimodal dialogue content due to a lack of reasoning and knowledge in the perception process.
\Cref{chatsearch} shows that ChatSearcher outperforms both CLIP and FROMAGe, suggesting that it possesses superior capability in comprehending image-text interleaved dialogues and discerning the implicit retrieval intentions effectively.

And we also find that for most models, performance on mChatSearch is higher than iChatSearch. 
Given the overlap in the last-round dialogue between two data parts, it indicates the necessity of incorporating external historical interaction context for better retrieval performance.

\subsection{Zero-shot Composed Image Retrieval}
\begin{table*}[t!]
\centering

\resizebox{\linewidth}{!}{

\begin{tabular}{l|l|cccc|ccc}
\toprule
 \multirow{2}{*}{Mode} & \multirow{2}{*}{Method}  & \multicolumn{4}{c|}{Recall@K }               & \multicolumn{3}{c}{R$_{subset}$@K }    \\
              &     &R@1&R@5&R@10&R@50 &R@1&R@2&R@3  \\
              \midrule
               & random choice &   0.04  & 0.22  &  0.44 &  2.18  &16.67&33.33&50.00            \\
\midrule
\multirow{3}{*}{Fine-tuned} &CIRPLANT~\cite{liu2021cirr} &   19.55   & 52.55  & 68.39  & 92.38    &    39.20&63.03&79.49       \\
&CompoDiff (T5-XL)~\cite{gu2023compodiff}&   22.35    & 54.36  &  \underline{73.41} &  \underline{91.77} &35.84&56.11&76.60        \\
&CLIP4Cir~\cite{clip4cir}       &   \textbf{38.53}   & \textbf{69.98} & \textbf{81.86} & \textbf{95.93} &\textbf{68.19}&\textbf{85.64}&\textbf{94.17}   \\
\midrule

\multirow{3}{*}{Zero-shot} & Pic2Word~\cite{saito2023pic2word}      &  23.90   & 51.70  & 65.30  &   87.80 &-&-&-                \\
  & CompoDiff (T5-XL)~\cite{gu2023compodiff}       &  19.37    & 53.81  &  72.02 &  90.85  &28.96&49.21&67.03 \\   
  & ChatSearcher          &  \underline{26.89}  & \underline{58.94}  &  72.68 & 91.42&\underline{43.61}&\underline{67.47}&\underline{80.43}        \\
\bottomrule
\end{tabular}

}
\label{cir}
\caption{Zero-shot composed image retrieval (CIR) results on CIRR test set. }
\end{table*}
Composed Image Retrieval (CIR) require the model to retrieve an image according to the reference image and user feedback text. 
We report the zero-shot retrieval performance of our model on a common used CIR benchmark CIRR~\cite{liu2021cirr} in ~\cref{cir}. 
Our model achieves state-of-the-art zero-shot performance on CIRR benchmark.
Meanwhile, ChatSearcher also outperform some fine-tuned methods as well. It shows the powerful transfer ability of our model to other image retrieval tasks.

\subsection{Zero-shot Text-based Image Retrieval}

\begin{table}[t!]
        \centering

\resizebox{0.85\linewidth}{!}{
\begin{tabular}{l|ccc|ccc}
\toprule
              & \multicolumn{3}{c|}{Flickr30K}             & \multicolumn{3}{c}{MSCOCO}                       \\
              & R@1&R@5&R@10 & R@1&R@5&R@10  \\
\midrule
CLIP~\cite{clip}        &     \textbf{68.7} & \textbf{90.6} & \textbf{95.2}    &     \underline{37.8} & \underline{62.4} & \underline{72.2}                \\

ChatSearcher (frozen LLM)            &  58.5      &    84.4&       90.3        &   33.7  &   59.6 & 70.5                              \\
ChatSearcher (w/o feature queue)            &  57.2      &    83.4&       89.3        &   31.7  &   58.1 & 69.2                              \\
ChatSearcher         &   \underline{68.0} & \underline{87.0} & \underline{92.2}              &  \textbf{41.7} & \textbf{67.5} & \textbf{76.9}                        \\

\bottomrule
\end{tabular}}
\label{t2i}
\caption{Zero-shot text-to-image retrieval results on Flickr30K and MSCOCO datasets. }
\end{table}

To validate the effectiveness of bidirectional image-text alignment training, 
we evaluate the model's zero-shot image retrieval capabilities after stage1 on two common text-to-image retrieval dataset MSCOCO~\cite{coco} and Flickr30K~\cite{flickr}. For evaluation, we use the Karpathy test split of MSCOCO and the test split of FLickr30K, comprising 5k and 1k images, respectively.  
As shown in \cref{t2i}, our model's performance is comparable with CLIP, indicating the successful establishment of image-text alignment. This underscores the significant potential of LLMs as multimodal context encoders.

\subsection{Visual Conversation Evaluation}
ChatSearcher is a model that integrates both image retrieval and visual dialogue capabilities. We have further fine-tuned ChatSearcher using the LLaVA665K\cite{liu2023improved} dataset, a widely used dataset for visual dialogue. We evaluated its performance on common visual conversation benchmarks. These benchmarks include vision question answering tasks: VQAv2\cite{vqa} and GQA\cite{gqa}. Furthermore, we tested ChatSearcher on multi-modal dialogue benchmarks, including MMBench~\cite{liu2023mmbench} and SEED-Bench~\cite{li2023seed}.
Our evaluation results indicate that the ChatSearcher achieves comparable performance to other state-of-the-art (SoTA) visual dialogue models, such as Qwen-VL~\cite{bai2023qwen} and LLaVA-1.5~\cite{liu2023improved}, demonstrating strong capabilities in handling visual dialogue tasks.
\begin{table*}[t!]
\centering

\resizebox{0.6\linewidth}{!}{
\begin{tabular}{l|cc|ccc}
\toprule
\bf Model  & \bf VQAv2 & \bf GQA  & \bf MMB & \bf SEED  \\
\midrule
BLIP-2~\cite{blip2}   &41.0 & 41.0  & -- & 46.4  \\
InstructBLIP-7B ~\cite{dai2023instructblip} & -- & 49.2 & 36.0  &  53.4          \\
IDEFICS-9B~\cite{laurenccon2023introducing} & 50.9 & 38.4  &  48.2 & --  \\
Qwen-VL~\cite{bai2023qwen}  &78.8 & 59.3  & 38.2  & 56.3     \\
Qwen-VL-Chat ~\cite{bai2023qwen}   &78.2 &57.5    & 60.6  & 58.2       \\
LLaVA-1.5-7B~\cite{liu2023improved}   & 78.5 & 62.0  & 64.3  &  \textbf{58.6}    \\
\midrule
ChatSearcher  & \textbf{78.9} & \textbf{62.5} & \textbf{64.7}  &  {58.1}   \\
\bottomrule
\end{tabular}}
\label{tab:visconv}
\caption{Comparison with SOTA models on visual conversation tasks. }
\end{table*}


\subsection{Ablation Study}


\paragraph{The choice of model design}
We mainly emphasize the importance of feature queue and trainable LLM. A common method to construct negative samples is to collect other image samples within a mini-batch. We compare this method with our feature queue method in \cref{t2i}. The feature queue provides more negative samples during the computation of feature matching probability, thereby enhancing the model's retrieval capability. We also find that the trainable LLM can effectively improve the retrieval performance in ~\cref{t2i}, suggesting that end-to-end training of LLM can improve the model’s capacity in comprehending the multimodal dialogue context information.

\begin{figure}
   \centering
        \includegraphics[width=0.7\linewidth]{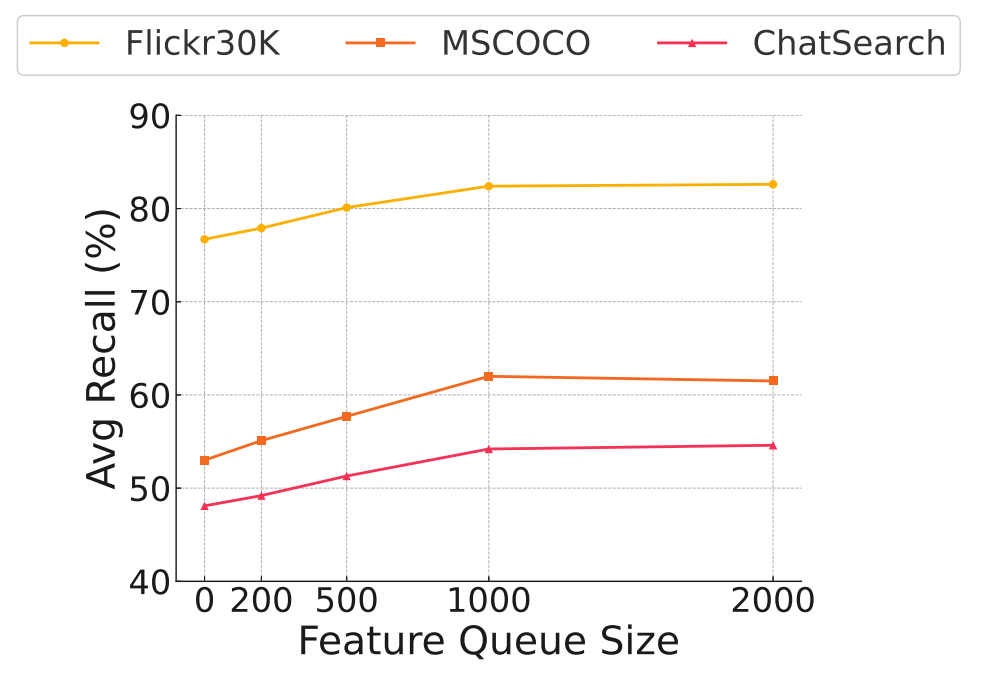}
        \caption{Ablation study on feature queue size. We show the average recall on three retrieval benchmarks: Flickr30k, MSCOCO and our ChatSearch.}
        \label{queuesize}
\end{figure}

\paragraph{The size of feature queue}
The feature queue is a crucial part of the ChatSearcher model. Negative samples from the feature queue play a key role in improving the model's ability to generate accurate query embeddings during contrastive learning. As shown in \cref{queuesize}, the model's performance in retrieval tasks gradually improves as the size of the feature queue increases. However, after the feature queue size reaches 1000, performance gains begin to plateau. Therefore, to balance storage requirements and performance, we selected 1000 as the final feature queue size.

\paragraph{The importance of visual conversation data}We add visual conversation data LLaVA-150k~\cite{llava} into our instruction data for two primary objectives: 1) enhancing the conversational capability of ChatSearcher for advanced reasoning and 2)bolstering the comprehension capacity of multimodal dialogue context for advanced retrieval. The first requirement is evidenced in conversational samples in \cref{ills} and the second requirement is manifested in the experimental results in \cref{abla}. By incorporating LLaVA-150k into our instructional dataset, we attain a superior performance compared to exclusively utilizing ChatSearch data for instruction tuning.

\begin{table*}[t!]
\centering

\resizebox{\linewidth}{!}{
\begin{tabular}{ccc|ccc|ccc|ccc|c}
\toprule
      \multicolumn{3}{c|}{Instruction Data}         & \multicolumn{3}{c|}{tChatSearch}   & \multicolumn{3}{c|}{iChatSearch}             & \multicolumn{3}{c|}{mChatSearch}   &   \multirow{2}{*}{Avg.}         \\
 ChatSearch &  LLaVA-150k  &  InstructPix2Pix   & R@1&R@5&R@10 & R@1&R@5&R@10 & R@1&R@5&R@10 & \\
\midrule
\checkmark & &      & 27.10  &52.00 &63.12 &34.86  & 61.02  &71.56   & 35.03  & 62.03 &  72.12  & 53.09\\
\checkmark & \checkmark &        & 27.64  & 52.54&63.38 & 35.13 & \textbf{61.20}  &  \textbf{71.59} & 35.59  & 62.51 & 72.11  & 53.52\\
\textcolor{gray}{\checkmark}$^\ast$& \checkmark & \checkmark   & \textbf{28.58}  & \textbf{52.86} &\textbf{63.82} & 29.25& 52.84  & 63.32  & 30.86  & 55.82 &  65.75 & 49.23\\
\checkmark&\checkmark&\checkmark     &  27.38 & 52.48& 63.50 & \textbf{35.54}  & 61.16 & 71.57& \textbf{37.90} & \textbf{64.22}  &  \textbf{74.06} & \textbf{54.20} \\
\bottomrule
\end{tabular}}
 
\label{abla}
\caption{Ablation study of the construction of instruction data. $^\ast$Gray \textcolor{gray}{\checkmark} indicates using only the 5k samples from tChatSearch training set. }
\end{table*}
\paragraph{The effect of adopting AIGC data} 
We find that AIGC data for image editing has following characteristics: high similarity between the reference and target images, user modification prompts similar with the feedback in a conversation. This suggests that AIGC data can be structured similarly to conversational image retrieval, 
which can serve as a valuable augmentation to instructional data. We randomly select 10k synthetic triplet samples from InstructPix2Pix~\cite{brooks2023instructpix2pix} dataset and incorporate them into the instruction dataset . The empirical results in ~\cref{abla} shows that these AIGC data can also improve model performance and enrich the diversity in our instructional data.
 \begin{figure}[t!]
   \centering
        \includegraphics[width=0.6\linewidth]{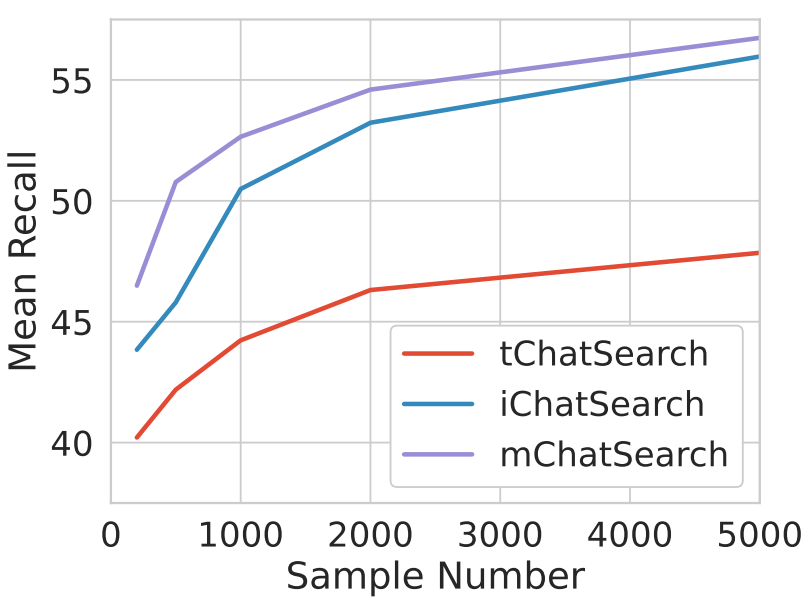}
        \caption{Ablation study on instruction data scale. }
        \label{scale}
\end{figure}
\paragraph{Instruction data scale}
In \cref{scale}, we explore the impact of scaling instruction data, while the average recall rate on rank 1, 5, 10 for each ChatSearch sub-task is reported. ChatSearcher can achieve better performance compared with CLIP~\cite{clip} with a limited number of instruction data, suggesting a strong multimodal contextual reasoning capability. Moreover, the retrieval performance is improved as the sample number increases, showing a notable scaling trend. 

\subsection{Qualitative Results}
\begin{figure*}[t!]
  \centering
     \includegraphics[width=\linewidth]{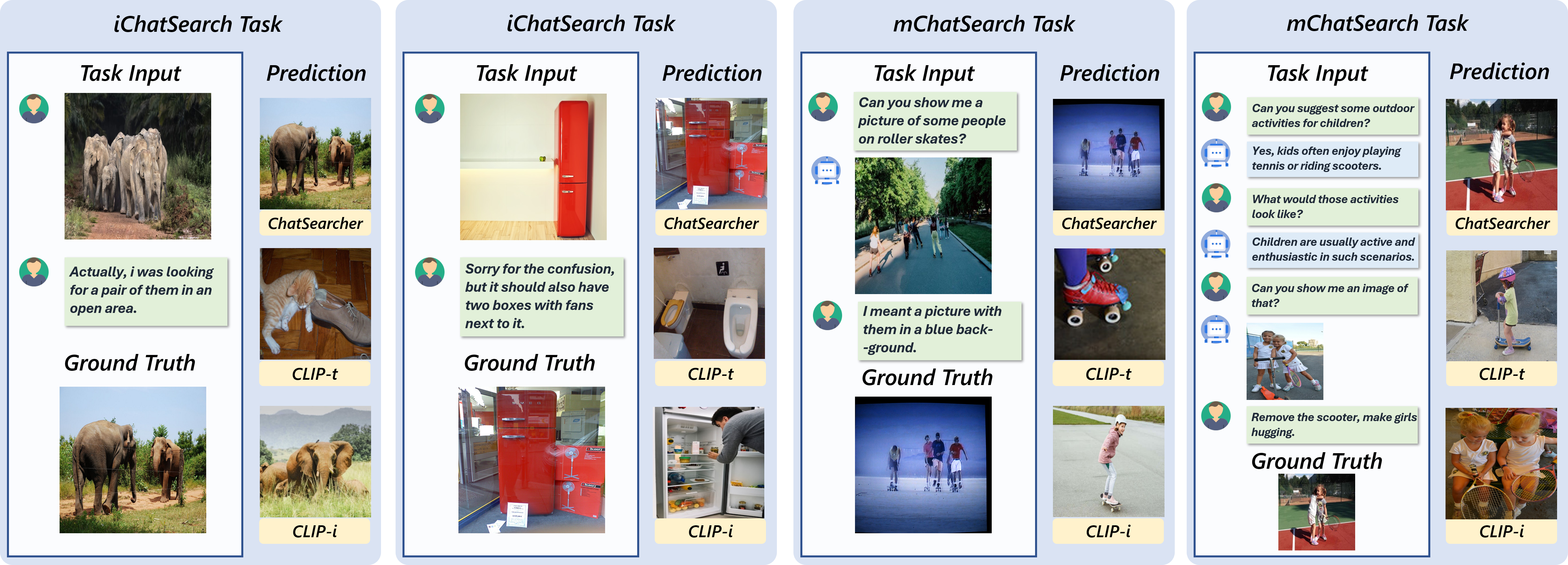}
     
  \caption{Qualitative results of ChatSearcher. We show ChatSearcher's conversational image retrieval ability across various dialogue contexts, exhibiting superiority over the vanilla CLIP approaches.
  }
\label{visualize}
\end{figure*}

\paragraph{Visualize on ChatSearch tasks}As depicted in \cref{visualize}, we provide a comparison among ChatSearcher and two retrievals method CLIP-t and CLIP-i mentioned in~\cref{clip-it} on iChatSearch task and mChatSearch task, which require the model to comprehend both visual and textual context concurrently. CLIP methods fail to handle the displayed cases, due to their restriction to understand explicit single-modality expressions. In contrast, ChatSearcher, enhanced by its multimodal reasoning capability, yields precise image retrieval results. 
\paragraph{Extension on grounding tasks} 
\begin{figure}[t!]
  \centering
     \includegraphics[width=0.8\linewidth]{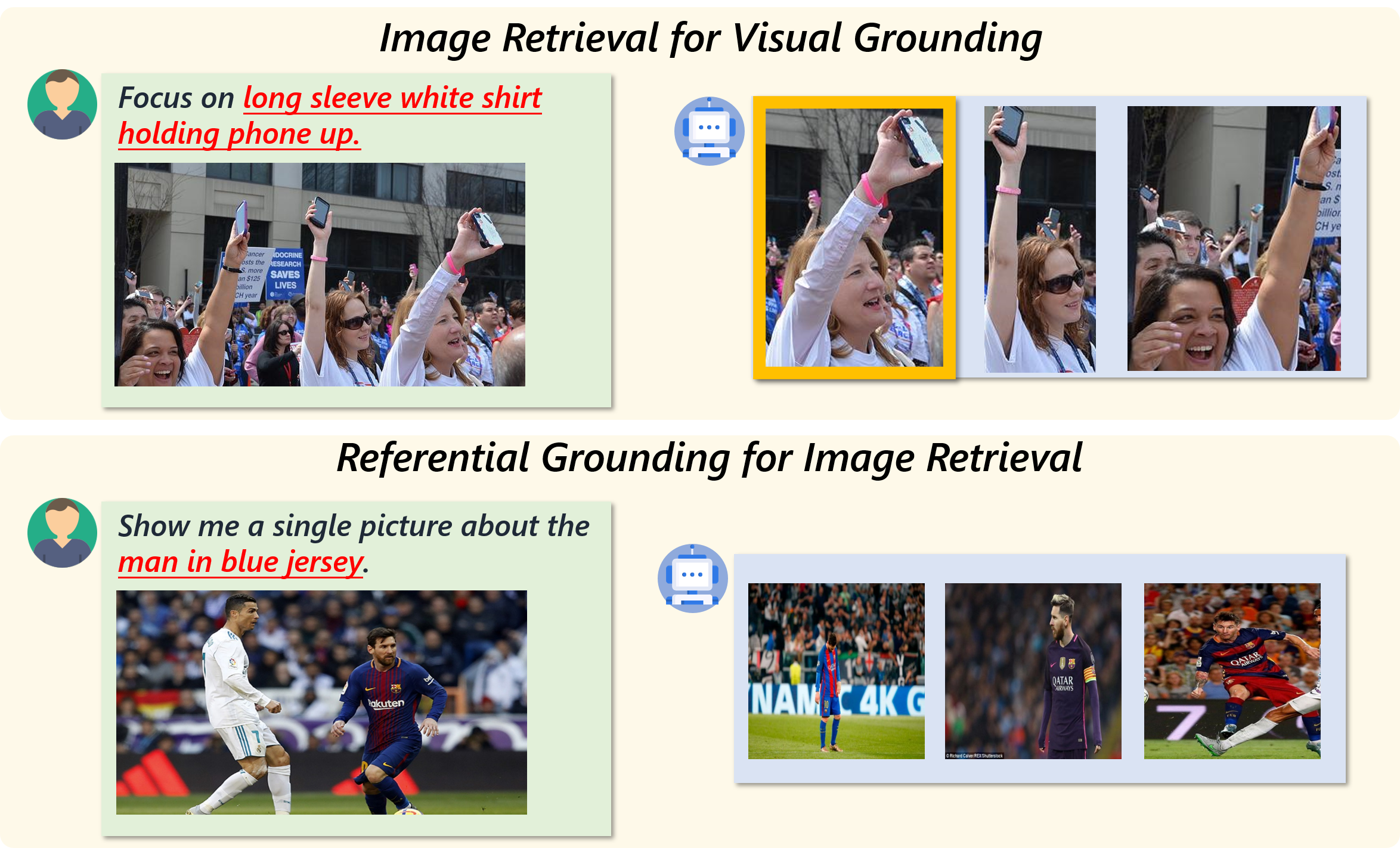}
  \caption{
 Qualitative results of combining grounding and retrieval: using retrieval result to find the region described by a textual phrase, and retrieving images based on a visual reference of the source image.
  }
\label{grounding}
\end{figure}
ChatSearcher has certain image-text grounding capabilities thanks to the alignment built in the first training stage, which is shown in \cref{grounding}. We explore two ways to combine grounding and retrieval: 1) using retrieval result for visual grounding: we use an offline object detector to extract bounding box proposals, crop them into single images,  subsequently retrieving results from above proposals according to the source image and reference descriptive text. 2)grounded reasoning to retrieval: we rely on the grounding capability of MLLM to retrieve for requests that consists visual reference, such as attributes or orientation, \etc. 

\begin{figure*}[t!]
  \centering

     \includegraphics[width=\linewidth]{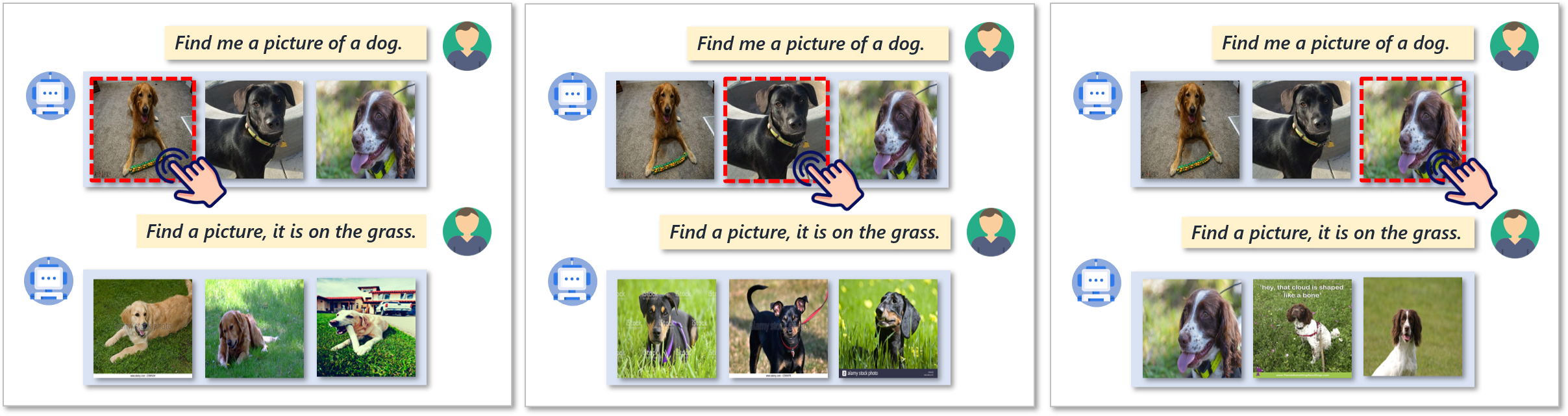}
  \caption{{Interaction on \textit{result choosing}. We show that \textit{different choices on image results} in previous round can influence the results in following round. In these samples, user choose different image returned by ChatSearcher and input same instruction to interact with model. ChatSearcher return different results based on user's choice and instruction.}}
  \label{branch1}
\end{figure*}

\begin{figure*}[t!]
  \centering

     \includegraphics[width=\linewidth]{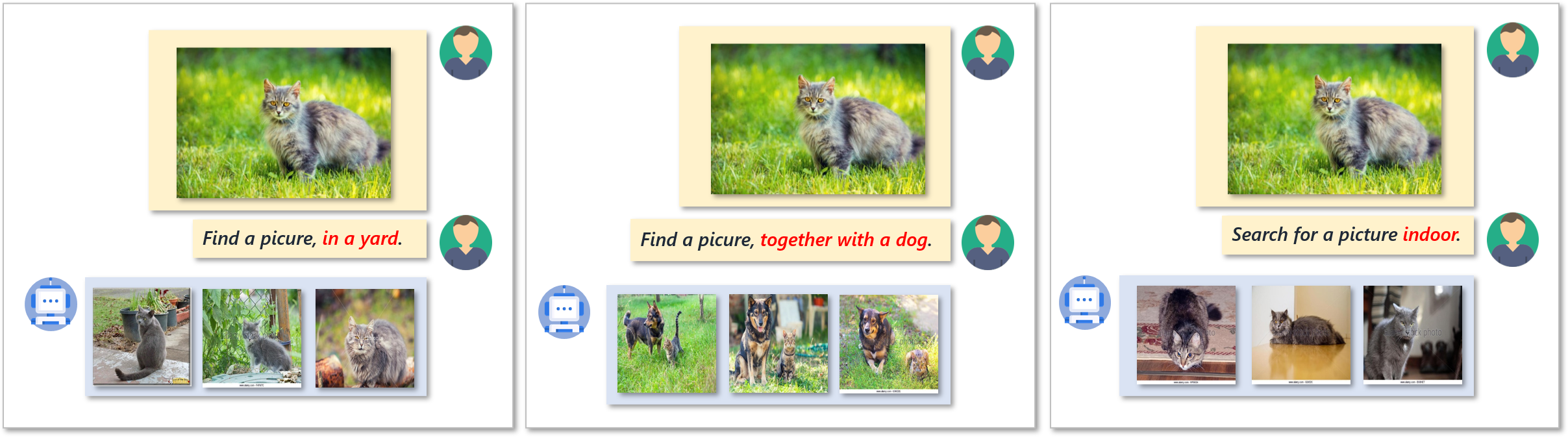}
       \vspace{-0.7cm}
  \caption{{Interaction on \textit{instruction choosing}. We show that \textit{different text instructions} on same image can influence the results. In these samples, user input different instructions with a same image. ChatSearcher return different results based on user's instruction and given image.}}
  \label{branch2}
   \vspace{-0.3cm}
\end{figure*}

\paragraph{Interactive inference.} We also show the interactive inference of ChatSearcher in \cref{branch1} and \cref{branch2}, illustrating how user choices can influence subsequent model inferences. In \cref{branch1}, we reveal that the image retrieval result selected by the user can impact the following results. Even when the same instructions are input during the second conversation round, the model can accurately identify the correct species based on the image context chosen by the user and provide different response. In \cref{branch2}, we further showcase how ChatSearcher can yield diverse image results by interpreting varying user instructions while provided with the same image input.

\section{Discussion and Conclusion}
Our dataset and model not only expand the frontiers of interactive image retrieval, but also enhance the essence of multimodal human-computer interactions. 
Our research offers a novel perspective on  enhancing multimodal outputs in human-computer conversations: presenting retrieved factual images to enhance the credibility and clarity of computer-generated information. Looking ahead, we intend to explore the extension of this fact-based credible output approach to diverse modalities including images, videos, audios, \etc.


In this paper, we have researched on the general conversational image retrieval task, aiming to extend image retrieval task into a more sophisticated interaction scenario where the retrieval intention is concealed within the multimodal dialogue context. To facilitate this study, we have curated a ChatSearch dataset using a meticulously designed automatic construction pipeline. Additionally, we propose a model called ChatSearcher, which operates under a generative paradigm to retrieve images by reasoning over their multimodal conversational context. 
ChatSearcher achieves outstanding performance on the general conversational image retrieval task and generalizes well to other zero-shot image retrieval tasks and visual conversation tasks. 
We anticipate that our work will provide novel perspectives in the fields of image retrieval and human-computer interaction.




\bibliographystyle{elsarticle-num}
\bibliography{refer}
\end{document}